\documentclass[10pt,twocolumn,letterpaper]{article}

\usepackage{iccv}
\usepackage{times}
\usepackage{epsfig}
\usepackage{graphicx}
\usepackage{amsmath}
\usepackage{amssymb}
\usepackage{multirow}
\usepackage{boldline}
\usepackage[ruled]{algorithm2e}

\DeclareMathOperator*{\argmin}{arg\,min}

\newcommand{\yq}{\textcolor{orange}}
\newcommand{\re}{\textcolor{black}}

\usepackage[pagebackref=true,breaklinks=true,letterpaper=true,colorlinks,bookmarks=false]{hyperref}

\iccvfinalcopy % *** Uncomment this line for the final submission

\def\iccvPaperID{622} % *** Enter the ICCV Paper ID here

\ificcvfinal\fi
\begin{document}
%%%%%%%%% TITLE & AUTHOR
\title{Self-similarity Grouping: A Simple Unsupervised Cross Domain Adaptation Approach for Person Re-identification}

% \author{Yang Fu\\
% IFP, Beckman Institute\\
% University of Illinois at Urbana-Champaign\\
% {\tt\small yangfu2@illinois.edu}
% % For a paper whose authors are all at the same institution,
% % omit the following lines up until the closing ``}''.
% % Additional authors and addresses can be added with ``\and'',
% % just like the second author.
% % To save space, use either the email address or home page, not both
% \and
% Thomas Huang\\
% IFP,Beckman Institute\\
% University of Illinois at Urbana-Champaign\\
% {\tt\small huang@ifp.uiuc.edu}
% }
\author{Yang Fu$^{1}$,  Yunchao Wei$^{1,2}$, Guanshuo Wang$^{3}$, Yuqian Zhou$^{1}$\\
Honghui Shi$^{4,1,5}$, Thomas S. Huang$^{1}$ \\ 
{\small $^1$University of Illinois at Urbana-Champaign},
{\small $^2$ReLER, University of Technology Sydney}, \\
{\small $^3$Shanghai Jiao Tong University},
{\small $^4$IBM Research},
{\small $^5$University of Oregon}\\
% {\tt\small \{yangfu2,yuqian2,t-huang1\}@illinois.edu, guanshuo.wang@sjtu.edu.cn}\\
% {\tt\small \{wychao1987,shihonghui3\}@gmail.com}
}

\renewcommand\footnotemark{}
\maketitle
%\thispagestyle{empty}

%%%%%%%%% ABSTRACT
\begin{abstract}
Domain adaptation in person re-identification (re-ID) has always been a challenging task. In this work, we explore how to harness the similar natural characteristics existing in the samples from the target domain for learning to conduct person re-ID in an unsupervised manner. Concretely, we propose a Self-similarity Grouping (SSG) approach, which exploits the potential similarity (from the global body to local parts) of unlabeled samples to build multiple clusters from different views automatically. These independent clusters are then assigned with labels, which serve as the pseudo identities to supervise the training process. We repeatedly and alternatively conduct such a \textbf{grouping} and \textbf{training} process until the model is stable. Despite the apparent simplify, our SSG outperforms the state-of-the-arts by more than 4.6\% (DukeMTMC$\rightarrow$Market1501) and 4.4\% (Market1501$\rightarrow$DukeMTMC) in mAP, respectively. Upon our SSG, we further introduce a clustering-guided semi-supervised approach named SSG$^{++}$ to conduct the one-shot domain adaption in an open set setting (\ie the number of independent identities from the target domain is unknown). Without spending much effort on labeling, our SSG$^{++}$ can further promote the mAP upon SSG by 10.7\% and 6.9\%, respectively. Our Code is available at: \href{https://github.com/OasisYang/SSG}{https://github.com/OasisYang/SSG} .
\end{abstract}
%%%%%%%%% BODY TEXT
\section{Introduction}\label{intro}
Person re-identification (re-ID) aims at matching images of a person in one camera with the images of this person from other different cameras. Because of its essential applications in security and surveillance, person re-ID has been drawing lots of attention from both academia and industry. Despite the dramatic performance improvement obtained by the convolutional neural network~\cite{hermans2017defense,sun2017beyond,wang2018learning}, it is reported that deep re-ID models trained on the source domain may have a significant performance drop on the target domain~\cite{deng2018image,fan2018unsupervised} due to the data-bias existing between source and target datasets. Since it is costly and unfeasible to annotate all images in target dataset, one of the most popular solutions for such issue is unsupervised domain adaptation (UDA). 

\begin{figure}[t]
	\centering
	\includegraphics[width=0.45\textwidth]{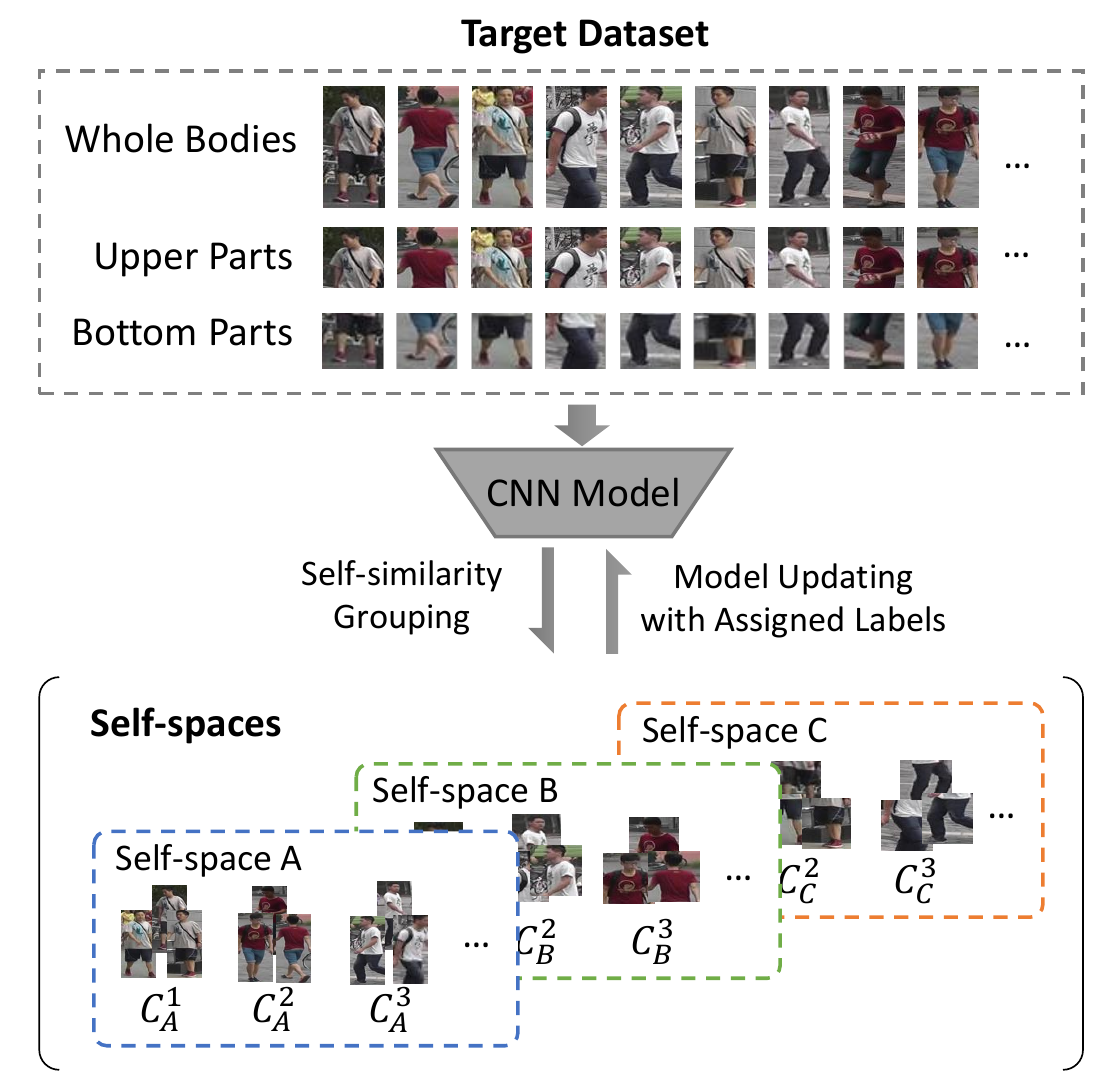}
	\caption{Illustration of proposed Self-similarity Grouping (SSG). We group target images by three cues, whole bodies, upper parts and bottom parts, independently and assign labels according to the corresponding group. The global information of the whole body and the local information of body parts can help us learn a better representation of individuals.}
	\label{fig:illus}
	\vspace{-3mm}
\end{figure}

Currently, the common UDA has been studied extensively in image classification, object detection, face recognition and semantic segmentation. However, the traditional UDA approaches~\cite{chen2018domain,chen2018road,motiian2017unified} always have an assumption that the source and target domain share the same set of classes, which does not hold for the person re-ID problem. Notably, in the person re-ID, different datasets have different identities (\ie classes). Recently, several unsupervised domain adaptation approaches for person re-ID have been proposed and achieve some promising improvements. Some works aim to translate the appearance of images from the source domain to the target domain based on the generative adversarial networks~\cite{deng2018image, wei2017person} by preserving the annotation information of the source domain. Besides, the disparities of cameras is another critical factor influencing re-ID performance, and HHL~\cite{zhong2018generalizing} is proposed to address intra-domain image variations caused by different camera configurations. However, the performances of these UDA approaches are still far behind their fully-supervised counterparts. The main reason is that most previous works focus on increasing the training samples or comparing the similarity or dissimilarity between the source dataset and the target dataset but ignoring the similar natural characteristics existing in the training samples from the target domain. %Actually, this kind of similarity is what we need for re-ID task. 

In order to address the problem above and discover the similarities among person images in target dataset, we proposed unsupervised Self-similarity Grouping (SSG) to mine the potential similarities from global to local manner. The critical idea of SSG arises from some recent re-ID works based on part matching, where different parts contain different discriminative information of a person. Fig~\ref{fig:illus} illustrates our proposed SSG approach. In particular, we extract features of all persons in target dataset and group them by three different cues, whole bodies (A), upper parts (B) and lower parts (C) independently. Then, we can obtain three sets of groups: ${C_A, C_B, C_C}$. By assigning a pseudo label to each group, we can pair every person with different pseudo labels. For instance, given a person $x_i$, it should be assigned by three pseudo labels, $C_A^i$, $C_B^i$ and $C_C^i$. As a result, we can establish a new dataset with pseudo labels, which can be treated as the normally labeled dataset. Since individuals with the same pseudo label should share lots of similarities, we iteratively mine these by finetuning the pre-trained model with the established dataset.

Upon our SSG, we further present a semi-supervised solution based clustering-guided annotation to approach the performance of the fully-supervised counterpart and efficiently achieve the adaption from the source domain to the target one. It is no more natural for us to think of the simplest semi-supervised solution, \ie one shot learning. In particular, the traditional one shot learning is based on the setting that only one sample from each category is labeled. However, the traditional one shot setting is not suitable for person re-ID case. Unlike image recognition problem, which is usually based on a closed set assumption, person re-ID problem is actually an open set problem. In other words, we cannot know in advance how many identities are included in a given unlabeled target dataset. Thus, the superior characteristics from traditional one shot setting cannot be directly applied to the re-ID case. To tackle the above-mentioned issue, we innovatively provide a clustering-guided semi-supervised solution.
% , which can simply yet effectively address the open set problem raised by re-ID. 
The proposed semi-supervised training strategy is based on the clustering-guided annotations, which samples a single image from each clustering. By doing this, we can significantly avoid choosing the same identity as two different ones. Hence, it allows us to leverage traditional one shot learning method and achieve similar performance.
We summarize our {\bf contributions} as follows:
\begin{itemize}
\item We propose the Self-similarity Grouping (SSG) method, which is a simple yet effective unsupervised domain adaptation (UDA) framework for person re-ID, in order to recover the performance of its fully supervised counterpart.
\item We introduce a similarity-guided semi-supervised training strategy for person re-ID and integrate it into the UDA framework, so that we can train unsupervised branch and semi-supervised branch jointly and effectively boost the process of domain adaption.
\item We conduct extensive experiments and ablation study on several popular benchmarks including Market1501~\cite{zheng2015scalable}, DukeMTMC-ReID~\cite{ristani2016performance,zheng2017unlabeled} and MSMT~\cite{wei2017person}, to demonstrate the effectiveness of proposed SSG and semi-supervised solution.
\end{itemize}
%------------------------------------------------------------------------
\section{Related Work}\label{related}
{\bf Unsupervised domain adaptation.}
Our work is closely related to unsupervised domain adaptation (UDA) where no data in target domain are labeled during training. Some works in this community try to address this problem by reducing the discrepancy between source domain and target domain~\cite{chu2016best,sun2016return,yosinski2014transferable}. For example, CORAL~\cite{sun2016return} learns a linear transformation that aligns the mean and covariance of feature distribution between two domains. And Sun~\cite{sun2016deep} proposes deep CORAL to extend original approach to deep neural networks with a nonlinear transformation. Some other methods aim to learn a transformation to generate samples that are similar to target domains by adversarial learning approach~\cite{bousmalis2017unsupervised,liu2016coupled,isola2017image}. Recently, some works solve this problem by mapping the source data and target data to the same feature space for the domain-invariant representations~\cite{ganin2014unsupervised, ganin2016domain,motiian2017unified,tzeng2015adapting}. For instance, Ganin~\etal~\cite{ganin2014unsupervised} propose a gradient reversal layer (GRL) and integrate it into a standard deep neural network for minimizing the classification loss while maximizing domain confusion loss. However, most of existing unsupervised domain adaptation methods are based on an assumption that class labels are the same across domains, while the person identities of different re-ID datasets are entirely different. Hence, the approaches mentioned above cannot be utilized directly for person re-ID task.
\begin{figure*}[t]
	\centering
	\includegraphics[width=0.95\textwidth]{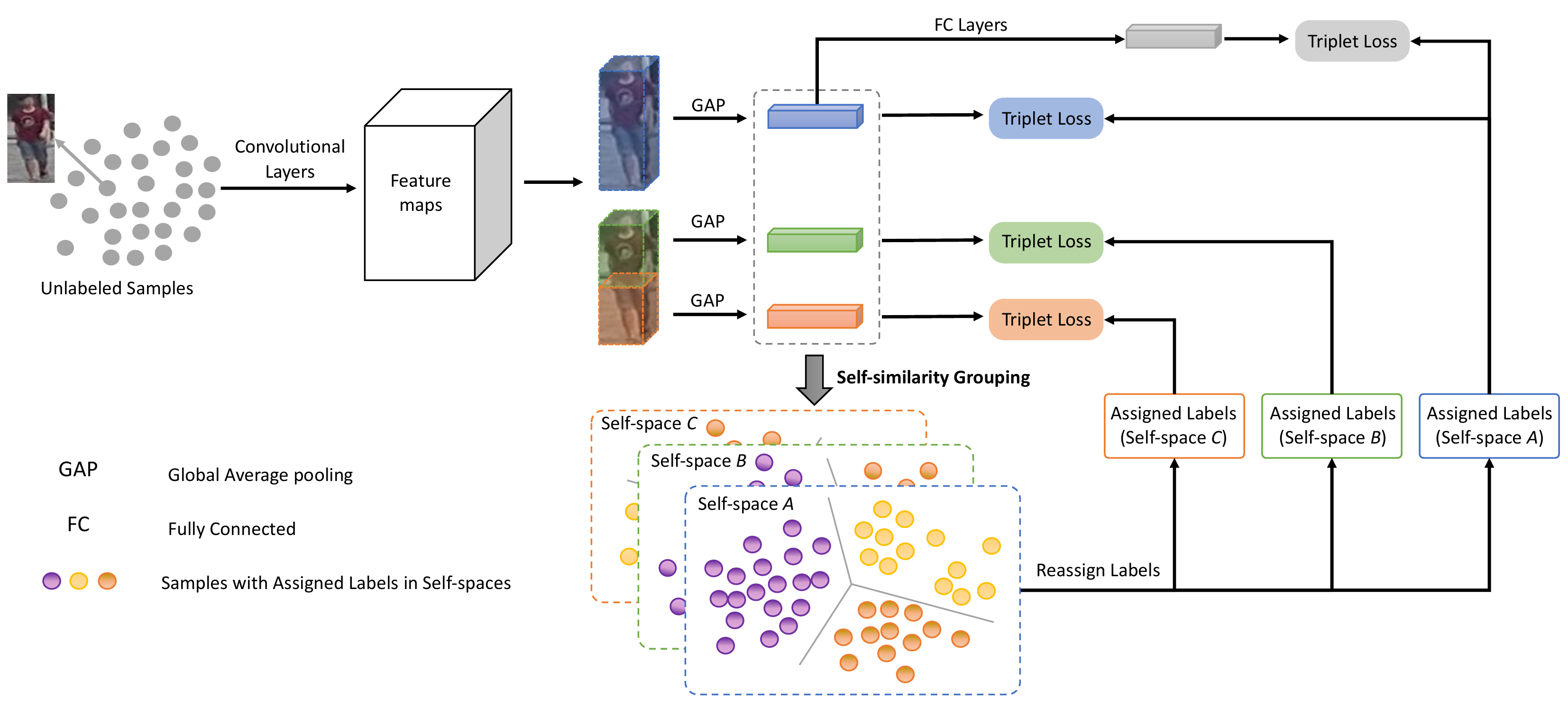}
	\caption{Overview of the proposed SSG approach. The CNN model is ResNet50 and pre-trained on source dataset. For each iteration, after feature extraction, we (1) split the feature maps into an upper part and a lower part and employ GAP on the whole, upper and lower feature maps. (2) Then, we group person images with different feature representations(blue, green and orange) and assign each different self-pseudo labels by grouping results(A, B, C). (3) Next, we update the CNN model by minimizing the triplet loss with each pseudo label. (4) During testing, we concatenate three feature representations together as the final representation of query person.}
	\label{fig:framework}
	\vspace{-3mm}
\end{figure*}

{\bf Unsupervised re-ID.} Some methods based on hand-craft features~\cite{bazzani2013symmetry,gray2008viewpoint,liao2015person} can be directly applied for unsupervised person re-ID. However, these methods always have a poor performance on large-scale dataset because they ignore the distribution of samples in the dataset. Benefit from the success of deep learning, some recent works~\cite{deng2018image,peng2016unsupervised, wang2018transferable, wei2017person} attempt to address unsupervised domain adaptation based on deep learning framework. Deng~\etal~\cite{deng2018image} aim to translate images from the source domain to the target domain by proposed similarity preserving generative adversarial network(SPGAN). And the translated images are utilized for training re-ID model in a supervised way. In~\cite{wang2018transferable}, a Transferable Joint Attribute-Identity Deep Learning (TJ-AIDL) is proposed to learn an attribute-semantic and identity discriminative feature representation space for target domain without using additional labeled data in the target domain. In~\cite{zhong2018generalizing}, Zhong~\etal introduce a Hetero-Homogeneous Learning (HHL) method, which aims to improve the generalization ability of re-ID models on the target set by achieving camera invariance and domain connectedness simultaneously. Although these unsupervised domain adaptation approaches achieve promising progress, the performance is still unsatisfactory compared with the fully supervised approaches.

{\bf Semi-supervised re-ID.} Semi-supervised learning aims at learning a task from one or very few training examples~\cite{fei2006one}, and there are some works of one shot person re-ID~\cite{bak2017one,figueira2013semi,liu2014semi, wu2018exploit}. In~\cite{bak2017one}, Bak~\etal utilize a metric learning approach for a pair of cameras which can be split into texture and color components for one shot image-based re-ID. Wu~\etal~\cite{wu2018exploit} propose a progressive sampling method to gradually predict reliable pseudo labels and update deep model for one shot video-based re-ID. However, the previous one shot re-ID works actually does not make sense. As described in the previous section, re-ID problem is an open set problem which means we cannot know how many identities in that dataset, so we cannot achieve the one/few shot setting under this situation. Based on the above analysis, in this paper, we aim to address person re-ID domain adaptation by Self-similarity grouping approach and further improve the performance by clustering-guided semi-supervised training.

%------------------------------------------------------------------------
\section{Proposed Method}
{\bf Problem Definition}
For unsupervised domain adaptation in person re-ID, we have a labeled source dataset $\{X_S, Y_S\}$, which contains $N_s$ person images and each image $x_s^i$ has a corresponding label $y_s^i$, where $y_s^i \in \{1, 2, ..., P_s\}$. $P_s$ is the number of identities in the source dataset. Also, we have another target dataset $\{X_t\}$, which consists of $N_t$ person images. Note that the identity of each image $x_t$ in target dataset $\{X_t\}$ is unknown. The goal of UDA person re-ID is to learn great discriminative embeddings of target dataset by only using the supervised information of source dataset.

\subsection{Fully Supervised Pre-training} \label{sec3.1}
Many existing UDA approaches are based on a model pre-trained on source dataset, and we follow the similar setting in~\cite{fu2019sta,zhong2017re,zhong2018generalizing,zhong2018camera} to obtain the pre-trained model. In particular, we first utilize ResNet50~\cite{he2016deep} pre-trained on ImageNet~\cite{deng2009imagenet} as backbone network. The last fully connected (FC) layer is discarded and two additional FC layers are added. The first one has 2048 dimensions named as ``FC-2048". The output of second FC layer is $P_s$ dimensional, where $P_s$ is the number of identity in source dataset, named as ``FC-\#ID". Given each labeled image $x_s$ in the source dataset and its ground truth identify $y_s$, we train the baseline model with cross-entropy loss and hard-batch triplet loss~\cite{hermans2017defense}. Specifically, the cross-entropy loss is employed with ``FC-\#ID" by casting the training process as a classiﬁcation problem and hard-batch triplet loss is employed with ``FC-2048" by treating the training process as a verification problem. We name this model as {\bf baseline} in this paper. The {\bf baseline} model achieves good performance with fully labeled data, but always fails when adopted to a new dataset.

\subsection{Unsupervised Self-similarity Grouping} \label{sec3.2}
Although the re-ID performance drops dramatically when directly adopted to another dataset, it is still much better than the performance of directly applying the ResNet50 pre-trained on ImageNet, which is almost zero. From this observation, we believe that the model trained on source dataset still learns some useful representations of a person for the re-ID task. The reason why it performs so severely on target dataset is that the similarities among different person images cannot be discovered correctly. In order to mine these similarities and make use of them for the re-ID task, we propose Self-similarity Grouping(SSG) approach. The overview of proposed SSG approach is shown in Fig~\ref{fig:framework}. The motivation of SSG is that we aim to encourage the model to discover the similarities existing in target dataset by self-similarity grouping. Then each unlabeled person is assigned with a pseudo label based on grouping results, which can be further used to reconstruct the target dataset and fine-tune the {\bf baseline} model. Inspired by recent re-ID work~\cite{fu2018horizontal, wang2018learning}, we compare the similarities between two persons not only by global information obtaining from the whole body but also by more fine and local information getting from upper and lower parts of a person. By combining the global and local information, we can obtain a more robust and discriminative representation of a person, which is more informative for self-similarity grouping.

To formulate the porposed SSG algorithm, we first feed each unlabeled images $x_t^i$ in target dataset into the {\bf baseline} model trained by configurations described in Sec~\ref{sec3.1} for feature extraction, denote as $F_t^i \in {\rm I\!R}^{H\times W\times C}$ %and the size of $F_t^i$ is $H\times W\times C$ 
(blue one in Fig~\ref{fig:framework}). Then, we split $F_t^i$ into two parts horizontally and each part contains the information of upper body or lower body and denoted as $F_{t\_up}^i \in  {\rm I\!R}^{\frac{H}{2} \times \frac{W}{2} \times C}$ (green one) and $F_{t\_low}^i \in  {\rm I\!R}^{\frac{H}{2} \times \frac{W}{2} \times C}$ (orange one).
%, whose size is $\frac{H}{2} \times \frac{W}{2} \times C$ (green one and orange one).
Next, we employ the Global Average Pooling (GAP) operation on whole feature map and two sliced feature maps, \ie $F_t^i$, $F_{t\_up}^i$ and $F_{t\_low}^i$, to obtain three feature vectors $f_t^i$, $f_{t\_up}^i$ and $f_{t\_low}^i$. We repeat above steps on every unlabeled images to generate three sets of feature vectors.
\begin{equation}
\left\{
\begin{aligned}\label{equ:fs}
    &f_t =  \{f_t^1,..., f_t^k,..., f_t^{N_t}\}\\
    &f_{t\_up}= \{f_{t\_up}^1,..., f_{t\_up}^k,..., f_{t\_up}^{N_t}\} \\
    &f_{t\_low} = \{f_{t\_low}^1,..., f_{t\_low}^k,..., f_{t\_low}^{N_t}\}
\end{aligned}
\right.
\end{equation} 
Based on these feature vectors sets, we utilize unsupervised clustering algorithm~\cite{ester1996density} on each set to obtain a series of groups, leading to every person image can be assigned with a pseudo label according to the group it belongs to, named as self-label. As shown in Fig~\ref{fig:framework}, we group images according to the three kinds of feature vectors, thus we can get three self-labels for each image $x_t^i$, denoted as $y_{t}^i$, $y_{t\_up}^i$ and $y_{t\_low}^i$. As a result, we can establish a new target dataset where each image has three self labels based on grouping result of three feature vectors, described as following.  
\begin{equation}
\begin{aligned}
    X_T & = \{x_t^i:(y_{t}^i, y_{t\_up}^i, y_{t\_low}^i); 1\leq i \leq N_t\}
\end{aligned}
\end{equation} 
In addition to the feature set in equation (\ref{equ:fs}), following the setting of the {\bf baseline} model, we also employ one FC layer after the $f_t^i$ to get a global embedding vector $f_{t\_e}^i$, which is 2048-dims and shares the same self label with $f_t^i$. Note that this FC layer will also be updated during training.

Finally, we employ the self-labels as the supervised information to fine-tune the pre-trained model for cross dataset adaptation using triplet loss $L_{triplet}$, which will be elaborated in Section~\ref{sec3.4}. Specifically, given an image, each feature vector and its corresponding self-label are used as two inputs for $L_{triplet}$. The full objective function of SSG is formulated as following, %where $L$ is a loss function used during training and more details will be explained in Section~\ref{sec3.4}.

% \yq{[comment: a little confused here. It's better to clarify which layers you will fix and which layer is fine-tuned. And whether you will fix FC2048 for all loss components? L is unclear here, maybe give the full definition of the combined loss.]}

\begin{equation}
\begin{aligned}
L_{ssg} = & L_{triple}(f_t, y_t) + L_{triple}(f_{t\_up}, y_{t\_up}) \\
& + L_{triple}(f_{t\_low}, y_{t\_low}) + L_{triple}(f_{t_e}, y_t) 
\end{aligned}
\end{equation} 
During training, we follow the above steps to fine-tine the {\bf baseline model} and employ the SSG for each iteration. During testing, we concatenate $f_t^i$, $f_{t\_up}^i$, $f_{t\_low}^i$ together as the final representation of each image $x_t^i$.

\begin{figure}[t]
	\centering
	\includegraphics[width=0.5\textwidth]{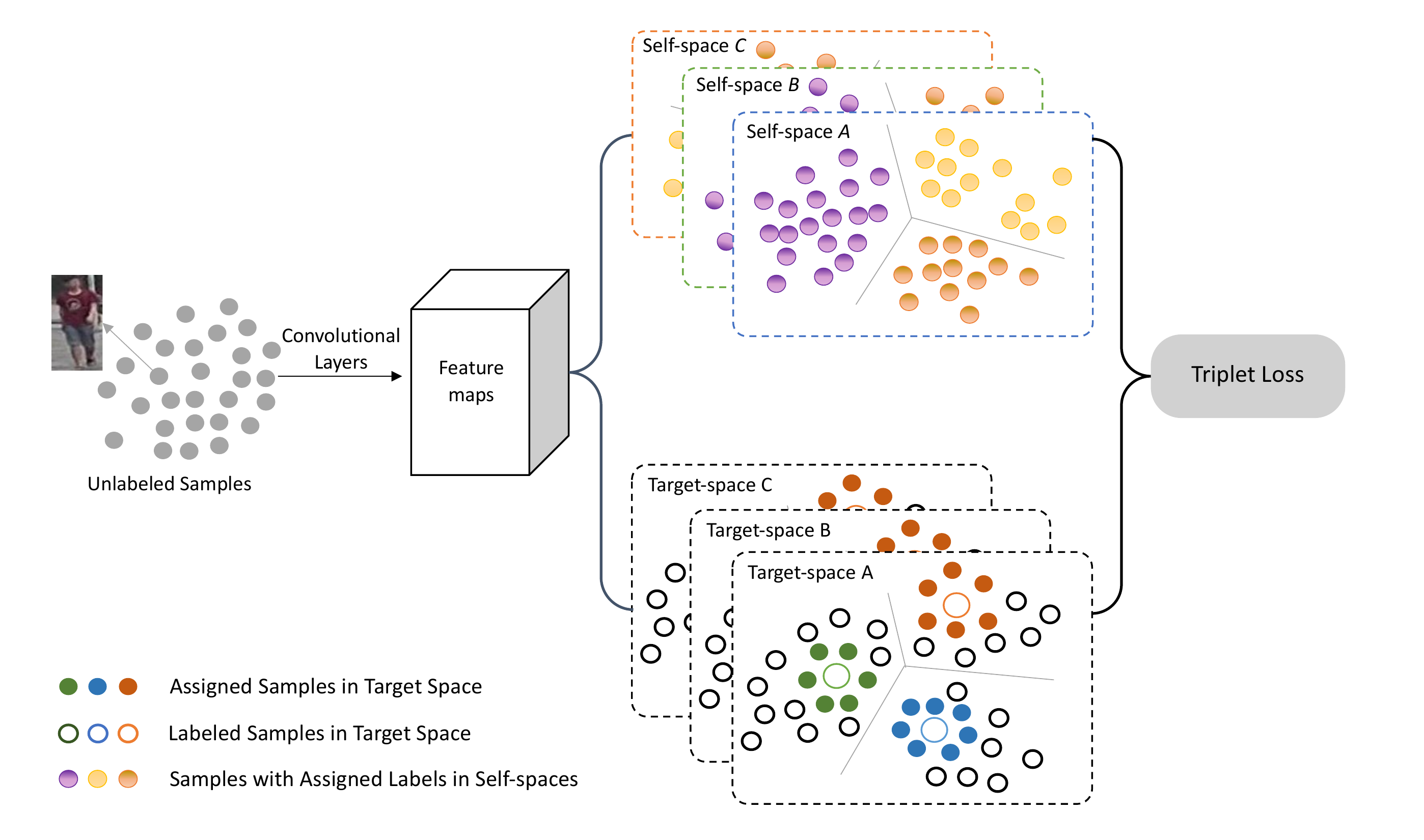}
	\caption{Overview of proposed clustering-guided semi-supervised training strategy combined with SSG. Given a target dataset, we annotate a few images based on the clustering results. And from these labeled images, we employ the step-wise learning approach to exploit the dataset gradually and obtain a more robust model.}
	\label{fig:semi}
	\vspace{-3mm}
\end{figure}

\subsection{Clustering-guided Semi-Supervised Training} \label{sec3.3}

Although unsupervised domain adaptation for person re-ID has been studied extensively~\cite{deng2018image,wei2017person,wang2018transferable,zhong2018generalizing}, there is still more than 25\% and 15\% performance drop in mAP and Rank-1 accuracy comparing to their fully supervised counterparts. In order to narrow the enormous performance gap, we further extend our SGG to a semi-supervised setting.

Previous semi-supervised person re-ID works are \re{mostly} under an one shot setting, which is actually not practical in the re-ID case since it is hard to know the number of independent identities in advance for a new dataset (as discussed in Section~\ref{related}). Hence, we introduce a new semi-supervised training strategy based on clustering-guided annotation, which is more practical and useful for real-world applications. As shown in Fig~\ref{fig:semi}, we employ the unsupervised clustering algorithm on $f_t$ to generate $N_g$ groups. Then, we randomly sample a single image from each group to form a very small sub-dataset $X_g$ with $N_g$ images. Next, we label this small sub-dataset manually and perform labels assignment based on this annotation. Specifically, we extract features of all images in sub-dataset $X_g$ and follow the same operation described in Section~\ref{sec3.2} to obtain three feature vector sets $f_g$, $f_{g\_up}$ and $f_{g\_low}$ and treat each of them as an identity dictionary. Given an unlabeled image $x_t^i$, we find the most similar images from $X_g$ by different cues, whole bodies, upper parts and lower parts, and assign $x_t^i$ with corresponding labels $y_{tg}^i$, $y_{tg\_up}^i$ and $y_{tg\_low}^i$, which can be formulated as following.
\begin{equation}
\left\{
\begin{aligned}
    & y_{tg}^i = \argmin_{k=1:N_g} dist\{f_t^i, f_g^k\} \\
    & y_{tg\_up}^i = \argmin_{k=1:N_g} dist\{f_{t\_up}^i, f_{g\_up}^k\} \\
    & y_{tg\_low}^i = \argmin_{k=1:N_g} dist\{f_{t\_low}^i, f_{g\_low}^k\}
\end{aligned}
\right.
\end{equation}

% Then, we randomly select one image from each group to form a very small sub-dataset $X_g$. Since these images are from different groups, it's less possible that two different images share the same identity. Then, we only need to annotate this sub-dataset with $N_g$ images, which is much cheaper and easier comparing to annotating the whole dataset. After that, we mainly follow the step-wised learning approach\cite{wu2018exploit} with some minor modifications. Specifically, . Given an unlabeled image $x_t^i$, we compare its similarity with every image in $X_g$ and assign it with the identity of the most similar one. Also, the same operation is performed on three feature vectors mentioned in Sec~\ref{sec3.2}, .
% and record the corresponding distance
% \begin{equation}
% \left\{
% \begin{aligned}
%     & y_{tg}^i = \argmin_{k=1:N_g} dist\{f_t^i, f_g^k\} \\
%     & y_{tg\_up}^i = \argmin_{k=1:N_g} dist\{f_{t\_up}^i, f_{g\_up}^k\} \\
%     & y_{tg\_low}^i = \argmin_{k=1:N_g} dist\{f_{t\_low}^i, f_{g\_low}^k\}
% \end{aligned}
% \right.
% \end{equation}
Note that we employ the k-reciprocal encoding~\cite{zhong2017re}, a variation of Jaccard distance between nearest neighbor sets, as the distance metric for similarity measurement.%where larger distance value means less similarity. 
Since each image in $X_g$ is from different groups, it's less possible that two different images share the same identity, which allows us to adopt some one shot learning approaches and further improve the performance. In particular, we follow the step-wise learning approach proposed in~\cite{wu2018exploit} to exploit the whole dataset during training stage progressively.

Furthermore, since SSG and clustering-guided semi-supervised training strategy share the same feature space, we can design a simple yet effective way to train the whole framework jointly and end-to-end, as shown in Fig~\ref{fig:semi}. And the superiority of joint training strategy will be illustrated in ablation study.

%And the proposed clustering-guided semi-supervised training exploits the whole dataset iteratively and we only assign some top images at each iteration and assign pseudo labels to all images in the end. Furthermore, since SSG and clustering-guided semi-supervised training strategy share the same feature space, we can design a simple yet effective way to train the whole framework jointly and end-to-end, as shown in Fig~\ref{fig:semi}. And the superiority of joint training strategy will be illustrated in ablation study.

\subsection{Loss Function} \label{sec3.4}
% \yq{[comment: I think it is better to merge the loss function section into the provious sections. ]}
\noindent {\bf Fully Supervised Training.} As describe in Sec~\ref{sec3.1}, we utilize both the batch-hard triplet loss proposed in \cite{hermans2017defense} and the softmax loss jointly.
The triplet loss with hard mining is first proposed in~\cite{hermans2017defense} as an improved version of the original semi-hard triplet loss~\cite{schroff2015facenet}. We randomly sample $P$ identities and $K$ instances for each mini-batch to meet the requirement of the batch-hard triplet loss. Typically, the loss function is formulated as follows:
\begin{equation}
\begin{aligned}
L_{triplet} = \sum_{i=1}^P\sum_{a=1}^K[\alpha + & 
\overbrace{\max_{p=1...K}||x_{a}^{(i)}-x_{p}^{(i)}||_2}^{hardest \ positive} \\ 
&- \underbrace{\min_{\substack{n=1...K\\
{j=1...P}\\
j \neq i}} ||x_{a}^{(i)}-x_{p}^{(i)}||_{2}}_{hardest \ negative}]_+
\end{aligned}
\end{equation}
where $x_{a}^{(i)}, x_{p}^{(i)}, x_{n}^{(i)}$ are features extracted from the anchor, positive and negative samples respectively, and $\alpha$ is the margin hyperparameter.
% to control the differences of intra and inter distances. Here, positive and negative samples refer to the person with same or different identity from the anchor.
Besides batch-hard triplet loss, we employ softmax cross entropy loss for discriminative learning as well, which can be formulated as follows:
\begin{equation}
    L_{softmax} = -\sum^P_{i=1}\sum^{K}_{a=1} \log \frac{e^{W^T_{y_{a, i}}x_{a,i}}}{\sum_{k=1}^{C}e^{W^T_{k}x_{a,i}}}
\end{equation}
where $y_{i,a}$ is the ground truth identity of the sample $\{a,i\}$, and $C$ is number of identity. Our loss function for optimization is the combination of softmax loss and batch-hard triplet loss as follows: 
\begin{equation}\label{eq:loss}
L_{baseline} = L_{softmax} + L_{triplet}
\end{equation}

\noindent {\bf Unsupervised and Semi-Supervised Training}
Unsupervised and semi-supervised training share the same loss function, and we just leverage the hard-batch triplet loss for metric learning. Also, each loss function has four components, whole body, upper body, lower body and global embedding, which can be formulated as follows:
\begin{equation}
\begin{aligned}
L_{ssg} = & L_{triple}(f_t, y_t) + L_{triple}(f_{t\_up}, y_{t\_up}) \\
& + L_{triple}(f_{t\_low}, y_{t\_low}) + L_{triple}(f_{t_e}, y_t) \\
L_{semi} = & L_{triple}(f_t, y_{tg}) + L_{triple}(f_{t\_up}, y_{tg\_up}) \\
& + L_{triple}(f_{t\_low}, y_{tg\_low}) + L_{triple}(f_{t_e}, y_{tg}) \\
\end{aligned}
\end{equation}
Hence, the objective function used in jointly training strategy is 
\begin{equation}\label{eq:oneshot}
L_{jointly} = L_{ssg} + L_{semi}.
\end{equation}

%------------------------------------------------------------------------
\section{Experiments}
In this section, we evaluate the proposed method on three large-scale re-ID datasets, % which are considered as large scale in the community, 
\ie Market1501~\cite{zheng2015scalable}\yq{,} DukeMTMC-ReID~\cite{ristani2016performance, zheng2017unlabeled} and MSMT17~\cite{wei2017person}. 

\subsection{Datasets and Evaluation Protocol}
{\bf Market1501}~\cite{zheng2015scalable} contains 32,668 images of 1,501 labeled persons \re{from} six camera views. Specifically, 12,936 person images \re{of} 751 identities detected by DPM~\cite{felzenszwalb2010object} are used for training. \re{For testing, in total} 19,732 person images \re{of} 750 identities plus some distractors form \re{the} gallery set, \re{and} 3,368 \re{manually cropped person regions} from 750 identities \re{form the} query set. %to retrieve the corresponding person images in the gallery set.

{\bf DukeMTMC-ReID}~\cite{zheng2017unlabeled} is a subset of the DukeMTMC dataset~\cite{ristani2016performance}. It contains 1,812 identities captured by 8 cameras. There are 16,522 training images, 2,228 query images, and 17,661 gallery images, with 1,404 identities appearing in more than two cameras. Also, similar \re{to} the Market1501, the rest 408 identities are considered as distractors. 
% \new{DukeMTMC-ReID is one of the most challenging re-ID datasets up to now with so many images from 8 multi-cameras.}

{\bf MSMT17}~\cite{wei2017person} is the largest re-ID dataset, which contains 126,441 bounding boxes of 4,101 identities taken by 15 cameras during 4 days. These 15 cameras \re{include} 12 outdoor and 3 indoor \re{ones}. Faster RCNN~\cite{ren2015faster} is utilized for pedestrian bounding box detection. \re{To the authors' best knowledge,} the MSMT17 is the most challenging re-ID dataset with \re{large-scale images and multiple cameras}.

{\bf Evaluation Protocol}
In our experiment, we use Cumulative Matching Characteristic (CMC) curve and the mean average precision (mAP) to evaluate the performance of re-ID. %CMC works 
% CMC represents the accuracy of the person retrieval, it is accurate when each query only has one ground truth. However, when multiple ground truths exist in the gallery, the goal is to return all right matches to the user. In this case, CMC may not have enough discriminative ability, but the mAP could reflect the recall. 
For Market-1501 and DukeMTMC-ReID, we use the evaluation packages provided by~\cite{zheng2015scalable} and~\cite{zheng2017unlabeled}, respectively. Moreover, for simplicity, all results reported in this paper are under the single-query setting, \re{and no post-processing like re-ranking ~\cite{zhong2017re} is applied.}%and do not use the re-ranking proposed in~\cite{zhong2017re} as post-processing. 
\subsection{Implementation Details}
{\bf Baseline training} As described in Section~\ref{sec3.1}, we first train a baseline model on \re{the} source dataset \re{by following} the training strategy described in~\cite{zhong2018camera}. Specifically, we keep the size of input images and resize them to $256\times128$. For data augmentation, we employ random cropping, flipping and random erasing~\cite{zhong2017random}. To meet the requirement of hard-batch triplet loss, each mini-batch is sampled with randomly selected $P=16$ identities and randomly sampled $K=8$ images for each identity from the training set, so that the mini-batch size is 128. And in our experiment, we set the margin parameter to 0.5. During training, we use the Adam ~\cite{kingma2014adam} with weight decay $0.0005$ to optimize the parameters for $150$ epochs. The initial learning rate is set to $3 \times 10^{-4}$ and decays to $3 \times 10^{-5}$ after 100 epochs.

{\bf Unsupervised and Semi-supervised training.} For unsupervised branch and semi-supervised branch, we follow the same data augmentation strategy and triplet loss setting. And we decrease the initial learning rate from $3 \times 10^{-4}$ to $6 \times 10^{-5}$ and change training epoch from 150 to 70. For fairness, we randomly select a single image from each clustering to annotate and preserve them for all ablation study. Besides, the whole framework is trained for several iterations until the model is stable.%\yq{[comment:why the last sentence?]}

Our model is implemented on Pytorch~\cite{paszke2017automatic} platform and trained with two NVIDIA TITAN X GPUs. All our experiments on different datasets follow the same settings as above.

\begin{table*}[t]\setlength{\tabcolsep}{12pt}
\centering
\footnotesize
\begin{tabular}{l|c|c|c|c|c|c|c|c}
\hlineB{2}
\multirow{2}{*}{Methods} & \multicolumn{4}{c}{DukeMTMC-Re-ID $\rightarrow$ Market1501 }  & \multicolumn{4}{|c}{Market1501 $\rightarrow$ DukeMTMC-ReID}  \\
\cline{2-9}
& mAP & R1 & R5 & R10 & mAP & R1 & R5 & R10  \\ \hline \hline
Baseline(Upper Bound) & 80.8 & 92.5 & 97.5 & 98.4 & 70.5 & 82.6 & 92.3 & 94.4\\ Direct Transfer & 26.6 & 54.6 & 71.1 & 77.1 & 16.1 & 30.5 & 45.0 & 51.8\\
UDA(paper)~\cite{song2018unsupervised} & 53.7 & 75.8 & 89.5 & 93.2 & 49.0 & 68.4 & 80.1 & 83.5 \\ \hline 
Baseline + UDA & 53.0 & 74.7 & 86.9 & 90.3 & 50.5 & 69.3 & 80.2 & 83.1\\
Baseline + SSG & 58.3 & 80.0 & 90.0 & 92.4 & 53.4 & 73.0 & 80.6 & 83.2\\
Baseline + SSG$^+$ & 62.5 & 81.4 & 91.6 & 93.8 & 56.7 & 74.2 & 83.5 & 86.7 \\
Baseline + SSG$^{++}$ & {\bf 68.7} & {\bf 86.2} & {\bf 94.6} & {\bf 96.5} & {\bf 60.3} & {\bf 76.0} & {\bf 85.8} & {\bf 89.3}\\ 
\hlineB{2}
\end{tabular}
\caption{Comparison of various methods on the target domains. When tested on DukeMTMC-reID, Market-1501 is used as the source and vice versa. ``Baseline" denotes using the full identity labels on the corresponding target dataset(See Section~\ref{sec3.1}). ``Direct Transfer" means directly applying the source-trained model on the target domain.``UDA" stands for the state-of-art unsupervised domain adaptation approach. ``Baseline+xxx" means using ``xxx" domain adaptation method upon baseline model. ``SSG" means Self-similarity Grouping in Section~\ref{sec3.2}. ``SSG$^+$" and ``SSG$^{++}$" stand for proposed SSG enhanced by clustering-guided semi-supervised training w/o and w/ joint training strategy described in Section~\ref{sec3.3}.}
% \vspace{-3mm}
\label{exp:t1}
\end{table*}

\subsection{Ablation Study}

{\bf Comparison between supervised learning, direct transfer and \re{state-of-the-art} unsupervised method.} The performance of supervised {\bf baseline} method and the direct transfer method are specified in Table~\ref{exp:t1}. When comparing two methods, we can clearly find that there is a large performance drop when directly adopting source-trained model on target dataset. For instance, the baseline model trained on Market1501 tested on Market1501 achieves 92.5\% in rank-1 accuracy and 80.8\% in mAP, but it drops to 30.5\% and 16.1\% when tested on DukeMTMC-reID, where the performance gap is more than 55\%. Also, a similar drop can be observed when DukeMTMC-reID is used as the training set and tested on Market1501. 
\re{Nevertheless, many recently proposed unsupervised domain adaptation methods cannot well address this great performance gap.} For example, when \re{transferring from} DukeMTMC-reID \re{to} Market1501, to our best knowledge, the best UDA approach~\cite{deng2018image} \re{only} achieved $53.7\%$ and $75.8\%$ on mAP and rank-1 accuracy, which is lower than the fully supervised method by about $20\%$.

{\bf Effectiveness of Self-similarity Grouping.} We perform several ablation studies to prove the effectiveness of proposed SSG as listed in Table~\ref{exp:t1}. Specifically, with SSG, we improve the performance by $21.2\%$ and $27.1\%$ in mAP and rank-1 accuracy when \re{the model is transferred from} DukeMTMC-reID \re{to} Market1501. \re{Similarly}, when \re{the model is} trained on Market1501 and tested on DukeMTMCreID, the performance gain \re{becomes} $+37.9\%$ and $+42.9\%$ in rank-1 accuracy and mAP, respectively. Moreover, compared with \re{state-of-the-art} UDA method, SSG can improve the performance by more than $5\%$ on Market1501 and more than $3\%$ on DukeMTMC-reID. This is because proposed SSG mines the potential similarity from global to fine manner, which can learn a more robust and discriminative model compared with~\cite{deng2018image}. Thus, SSG is a simple yet effective unsupervised method for domain adaptation in person re-ID. 
% \yq{[comment:if possible, briefly analyze why the performance is better, in terms of the method design and comparison.]}

In addition, \re{we compare the performance of SSG with different number of \re{sliced} horizontal feature spatial parts.}%as described in Section~\ref{sec3.2}, we slice the feature map into different spatial parts horizontally and group each of them. Hence, 
 \re{Table~\ref{exp:t2} suggests} that SSG achieves the best results when \re{we only split the feature maps into two parts: the upper and lower body.} 
 %grouping by whole body, upper part and lower part. This is because 
 \re{It infers that} the upper and lower body contain the most \re{discriminative} information for re-ID. %By slicing into more parts, some parts may not have enough information for similarity mining.
 \re{More fragments of features may break this information and yield worse similarity mining and matching.} 

\begin{table}\setlength{\tabcolsep}{11pt}
\centering
\footnotesize
\begin{tabular} {l|c|c|c|c}
\hlineB{2}
Methods & mAP & R1 & R5 & R10 \\ \hline 
UDA & 49.0 & 68.4 & 80.1 & 83.5 \\ \hline
SSG(2 parts) & {\bf 53.4} & {\bf 73.0} & {\bf 80.6} & {\bf 83.2} \\ 
SSG(3 parts) & 50.4 & 71.3 & 79.1 & 82.0 \\ 
SSG(4 parts) & 49.1 & 69.1 & 78.4 & 81.6 \\ 
\hlineB{2}
\end{tabular}
\caption{The performance of proposed SSG with different number of splitted parts when trained DukeMTMC-reID dataset and tested on Market1501 dataset. }
\label{exp:t2}
\vspace{-3mm}
\end{table}

% \begin{figure}[t]
% 	\centering
% 	\includegraphics[width=0.48\textwidth]{fig/compare_stepwise.pdf}
% 	\caption{Performance Comparison of step-wised one shot learning and similarity-guided one shot learning over training iterations on Market1501.}
% 	\label{exp:compare}
% 	\vspace{-5mm}
% \end{figure}

{\bf Effectiveness of clustering-guided semi-supervised training.} 
%We conduct several experiments to verify the influence of cluster-guided semi-supervised training on person re-ID domain adaptation, as shown in Table~\ref{exp:t1}. 
\re{Table~\ref{exp:t1} shows the performance of the proposed cluster-guided semi-supervised method (SSG$^{+}$). Compared to the unsupervised one (SSG), SSG$^{+}$ outperforms SSG by $4.2\%$ and $1.4\%$ in terms of mAP and rank-1 accuracy when tested on Market1501. In addition, it outperforms the direct transfer method by $35.9\%$ in mAP and $26.8\%$ in rank-1 accuracy. These demonstrate the effectiveness of the clustering-guided semi-supervised training.}
%First, we randomly select one image from each clustering to set up a very small sub-dataset. Then, we employ the semi-supervised training with the selected sub-dataset on model trained by SSG. Specifically, we exploits the target dataset gradually and assign the pseudo labels with highest confidence score step by step as described in~\cite{wu2018exploit}. And the number of selected pseudo labels increases iteratively.
%Comparing to unsupervised SSG, we improve mAP and rank-1 accuracy by $+4.2\%$ and $+1.4\%$ when test on Market1501. In addition, we gain $+38.8\%$ and $+30.8\%$ on mAP and rank-1 accuracy.

\begin{table}[t]\setlength{\tabcolsep}{12pt}
\centering
\footnotesize
\begin{tabular}{l|c|c|c}
\hlineB{2}
\multirow{2}{*}{Methods} & \multicolumn{3}{c} {DukeMTMC-reID$\rightarrow$ Market1501}  \\ 
\cline{2-4}
& mAP & R1 & R10 \\ \hline
SSG$^*$ & 66.8 & 84.5 & 95.3 \\ 
SSG$^{++}$ & 68.7 & 86.2 & 96.5 \\ \hline \hline
\multirow{2}{*}{Methods} & \multicolumn{3}{c} {Market1501$\rightarrow$ DukeMTMC-reID}  \\ 
\cline{2-4}
& mAP & R1 & R10 \\ \hline
SSG${^*}$ & 57.6 & 74.4 & 88.1 \\
SSG$^{++}$& 60.3 & 76.0 & 89.3 \\ \hline
\hlineB{2}
\end{tabular}
\caption{Comparison of proposed cluster-guided annotation with random sampling annotation on Market1501 dataset and DukeMTMC-reID dataset. SSG$^*$ is semi-supervised training with random sampling annotation}
\vspace{-3mm}
\label{exp:t3}
\end{table}

% \begin{figure}
% 	\centering
% 	\includegraphics[width=0.45\textwidth]{fig/pseudo_accuracy.png}
% 	\caption{Pseudo labels prediction accuracy comparison of clustering-guided annotation and random annotation over training iterations on Markett1501 dataset}
% 	\label{fig:pseudo}
% 	\vspace{-5mm}
% \end{figure}

{\bf Effectiveness of joint training strategy} 
\re{Instead of fine-tuning the model obtained by SSG for SSG$^+$, we conduct experiments to jointly train the model parameters using both SSG and SSG$^+$ losses, and we denote this jointly trained model as SSG$^{++}$. In Table~\ref{exp:t1}, when transferred from DukeMTMC-reID to Market1501, SSG$^{++}$ outperforms SSG$^+$ by $6.2\%$ and $4.8\%$ in terms of mAP and rank-1 accuracy. The performance of testing on DukeMTMC-reID also boosts by $3.6\%$ and $1.8\%$ in mAP and rank-1 accuracy. It indicates that joint training strategy has its superiority on both two datasets, and training also becomes more efficient.}
%s described in Section~\ref{sec3.2}, we can further combine clustering-guided semi-supervised training and SSG together and train them jointly, but not finetune on the top of model by obtained by SSG. %From Table~\ref{exp:t1}, we gain $+6.2\%$ and $+4.8\%$ in mAP and rank-1 accuracy ,respectively, when trained on DukeMTMC-reID and tested on Market1501. When tested on DukeMTMC-reID, the gains are $+3.6\%$ and $+1.8\%$ in mAP and rank-1 accuracy, respectively.By jointly training strategy, we can not only improve the reID performance on both dataset, but also save the training time.

{\bf Effectiveness of clustering-guided annotation }
\re{Intuitively, compared to random sampling from the target domain, clustering-guided annotation (\ie sampling from unsupervised clustered groups and annotating) will increase the identity diversity in the sample set, and enhance the learned feature representation ability with limited supervised information. To validate this intuition, we compare clustering-guided and random sampling annotation on the jointly learned SSG$^{++}$. For fairness, we randomly sampled the same number of images from the whole dataset for the latter. In Table~\ref{exp:t3}, SSG$^{++}$ surpasses the random sampling one by 2.7\% and 1.6\% in mAP and rank-1 accuracy when testing on DukeMTMC-reID and the similar improvement on performance can be found when testing on Market1501 as well. It verifies that the proposed clustering-guided annotation is better than the random one.} 

\re{In conclusion, SSG$^{++}$ with clustering-guided annotation yields the best performance on both Market1501 and DukeMTMC-reID dataset. For instance, we achieve $42.1\%$ and $31.6\%$ improvements in mAP and rank-1 accuracy when testing on Market1501 compared with direct adaptation.}
%In order to validate the effectiveness of clustering-guided annotation, we also compare semi-supervised training with cluster-guided annotation and with random sampling annotation. Note that by random sampling, we select the same number of images from the whole dataset randomly for fairness. The results are shown in Table~\ref{exp:t3}. It's obviously that the cluster-guided annotation method has much better performance than randomly sampling. 
% Also, Figure~\ref{fig:pseudo} illustrates the accuracy of pseudo label prediction semi-supervised training with proposed clustering-guided annotation and random annotation over training iterations. It's clearly that the accuracy of clustering-guided annotation(blue line) is much higher random annotation(red line). 

%This is because, comparing to random sampling, sampling from each clustering can significant lower the probability of choosing the same identities as two different ones, which allows us to learn more robust representation of target domain dataset with very limited supervised information.
%So far all components in our proposed person re-ID cross domain adaptation framework have been evaluated and validated, we achieve promising performances on both Market1501 and DukeMTMC-reID dataset. For instance, we achieve $45.0\%$ and $35.6\%$ improvements in mAP and rank-1 accuracy when trained on DukeMTMC-reID and tested on Market1501 compared with directly adaptation.

\subsection{Comparision with State-of-arts}
In this section, we compare the proposed method with \re{state-of-the-art} unsupervised learning methods on Market1501, DukeMTMC-reID and MSMT17 in Table~\ref{exp:t4}, Table~\ref{exp:t5} and Table~\ref{exp:t6} respectively. SSG outperforms existing approaches with dominantly advantage. In particular, our model outperforms the best published method ARN~\cite{li2018adaptation} by almost $20\%$ on mAP when testing on Market1501 and DukeMTMC-reID dataset. Moreover, it also surpasses the unpublished UDAP~\cite{deng2018image} and MAR(CVPR2019)~\cite{yu2019unsupervised} by a large margin.

{\bf Results on Market1501}
On Market-1501, we compare our results with two hand-crafted features, \ie Bag-of-Words (BoW)~\cite{zheng2015scalable} and local maximal occurrence (LOMO)~\cite{liao2015person}, three unsupervised methods, including UMDL~\cite{peng2016unsupervised},PUL~\cite{fan2018unsupervised} and CAMEL~\cite{yu2017cross}, and six unsupervised domain adaptation methods, including PTGAN~\cite{wei2017person}, SPGAN~\cite{deng2018image}, TJ-AIDL~\cite{wang2018transferable}, ARN~\cite{li2018adaptation}, UDAP~\cite{song2018unsupervised} and MAR~\cite{yu2019unsupervised}. The two hand-crafted features are directly applied to the test dataset without any training process, but it is obvious that both features fail to obtain competitive results. \re{While} training on target set, unsupervised methods always obtain higher results than hand-crafted features. Comparing with unsupervised domain adaptation methods, our proposed SSG is superior. Only \re{in} unsupervised setting, we achieve rank-1 accuracy $=80.0\%$ and mAP $=58.3\%$, which outperforms the best unsupervised method~\cite{song2018unsupervised} by $4.6\%$ and $4.2\%$. Furthermore, with clustering-guided semi-supervised training strategy, we \re{further} improve the performance by $10\%$ on mAP and $6\%$ on rank-1 accuracy.

\begin{table}\setlength{\tabcolsep}{11pt}
\centering
\footnotesize
\begin{tabular} {l|c|c|c|c}
\hlineB{2}
Methods & mAP & R1 & R5 & R10 \\ \hline \hline
LOMO~\cite{liao2015person} & 8.0 & 27.2 & 41.6 & 49.1 \\
Bow~\cite{zheng2015scalable} & 14.8 & 35.8 & 52.4 & 60.3 \\
UMDL~\cite{peng2016unsupervised} & 12.4 &34.5 & 52.6 & 59.6 \\
PTGAN~\cite{wei2017person} & - & 38.6 & - & 66.1 \\
PUL~\cite{fan2018unsupervised} & 20.5 & 45.5 & 60.7 & 66.7 \\
SPGAN~\cite{deng2018image}& 22.8 & 51.5 & 70.1 & 76.8 \\
CAMEL~\cite{yu2017cross} & 26.3 & 54.5 & - & - \\
SPGAN+LMP~\cite{deng2018image} & 26.7 & 57.7 & 75.8 & 82.4 \\
TJ-AIDL~\cite{wang2018transferable} & 26.5 & 58.2 & 74.8 & 81.1 \\
HHL~\cite{zhong2018generalizing} & 31.4 & 62.2 & 78.8 & 84.0 \\
ARN~\cite{li2018adaptation} & 39.4 & 70.3 & 80.4 & 86.3 \\
UDAP~\cite{song2018unsupervised} & 53.7 & 75.8 & 89.5 & 93.2 \\ 
MAR~\cite{yu2019unsupervised} & 40.0 & 67.7 & 81.9 & - \\ 
ENC~\cite{zhong2019invariance} & 43.0 & 75.1 & 87.6 & 91.6\\ \hline
SSG & {\bf 58.3} & {\bf 80.0} & {\bf 90.0} & {\bf 92.4} \\ 
SSG$^{++}$ & {\bf 68.7} & {\bf 86.2} & {\bf 94.6} & {\bf 96.5}\\ 
\hlineB{2}
\end{tabular}
\caption{Comparison of proposed SSG approach with state-of-arts unsupervised domain adaptive person re-ID methods on Market1501 dataset. }
\vspace{-3mm}
\label{exp:t4}
\end{table}

{\bf Results on DukeMTMC-reID}
The similar improvement can also be observed when we \re{test it} on DukeMTMC-reID dataset. Specifically, we achieve mAP $=53.4\%$ and rank-1 accuracy $=73.0\%$ by unsupervised SSG and obtain mAP $=60.3\%$ and rank-1 accuracy$=76.0\%$ under semi-supervised setting. Compared with best unsupervised method, our result is $4.4\%/11.3\%$ and $4.6\%/7.6\%$ higher on mAP and rank-1 accuracy. Therefore, the superiority of the proposed SSG adaptation approach for person re-ID can be concluded. In addition, we improve the performance by a large margin and recover about $90\%$ performance of the fully supervised method by clustering-guided semi-supervised training strategy.

\begin{table}\setlength{\tabcolsep}{11pt}
\centering
\footnotesize
\begin{tabular} {l|c|c|c|c}
\hlineB{2}
Methods & mAP & R1 & R5 & R10 \\ \hline \hline
LOMO~\cite{liao2015person} & 4.8 & 12.3 & 21.3 & 26.6 \\
Bow~\cite{zheng2015scalable} & 8.3 & 17.1 & 28.8 & 34.9 \\
UMDL~\cite{peng2016unsupervised} & 7.3 & 18.5 & 31.4 & 37.4 \\
PTGAN~\cite{wei2017person} & - & 27.4 & - & 50.7 \\
PUL~\cite{fan2018unsupervised} & 16.4 & 30.0 & 43.4 & 48.5  \\
SPGAN~\cite{deng2018image}& 22.3 & 41.1 & 56.6 & 63.0\\
CAMEL~\cite{yu2017cross} &  - & - & - & -\\
SPGAN+LMP~\cite{deng2018image} & 26.2 & 46.4 & 62.3 & 68.0 \\
TJ-AIDL~\cite{wang2018transferable} & 23.0 & 44.3 & 59.6 & 65.0 \\
HHL~\cite{zhong2018generalizing} & 27.2 & 46.9 & 61.0 & 66.7 \\
ARN~\cite{li2018adaptation} & 33.4 & 60.2 & 73.9 & 79.5 \\
UDAP~\cite{song2018unsupervised} & 49.0 & 68.4 & 80.1 & 83.5 \\ 
MAR~\cite{yu2019unsupervised} & 48.0 & 67.1 & 79.8 & - \\
ENC~\cite{zhong2019invariance} &40.4 &63.3 &75.8& 80.4\\ \hline
SSG & {\bf 53.4} & {\bf 73.0} & {\bf 80.6} & {\bf 83.2} \\ 
SSG$^{++}$ & {\bf 60.3} & {\bf 76.0} & {\bf 85.8} & {\bf 89.3} \\ 
\hlineB{2}
\end{tabular}
\caption{Comparison of proposed SSG approach with state-of-arts unsupervised domain adaptive person re-ID methods on DukeMTMC dataset. }
\vspace{-3mm}
\label{exp:t5}
\end{table}

{\bf Results on MSMT17}
In addition, we further evaluate the proposed SSG approach on MSMT17 dataset, which is the largest and most challenging re-ID dataset. We achieve mAP$=13.3\%$ and rank-1 accuracy$=32.2\%$ when trained DukeMTMC-reID, which improves state-of-the-arts by $10.0\%$ and $20.4\%$. Also, similar improvement can be observed while trained on Market1501 as well, which further verifies the effectiveness of our proposed method.

\begin{table}[t]\setlength{\tabcolsep}{12pt}
\centering
\footnotesize
\begin{tabular}{l|c|c|c}
\hlineB{2}
\multirow{2}{*}{Methods} & \multicolumn{3}{c} {DukeMTMC-reID$\rightarrow$ MSMT17}  \\ 
\cline{2-4}
& mAP & R1 & R10 \\ \hline
PTGAN~\cite{wei2017person} & 3.3 & 11.8 & 27.4 \\ \hline
SSG & {\bf 13.3} & {\bf 32.2} & {\bf 51.2} \\
SSG$^{++}$ & {\bf 18.3} & {\bf 41.6} & {\bf 62.2} \\ \hline \hline
\multirow{2}{*}{Methods} & \multicolumn{3}{c} {Market1501$\rightarrow$ MSMT17}  \\ 
\cline{2-4}
& mAP & R1 & R10 \\ \hline
PTGAN~\cite{wei2017person} & 2.9 & 10.2 & 24.4 \\ \hline
SSG & {\bf 13.2} & {\bf 31.6} & {\bf 49.6} \\ 
SGG$^{++}$ & {\bf 16.6} & {\bf 37.6} & {\bf 57.2}  \\ 
\hlineB{2}
\end{tabular}
\caption{Comparison of proposed SSG approach with state-of-arts unsupervised domain adaptive person re-ID methods on MSMT17 dataset.}
\vspace{-3mm}
\label{exp:t6}
\end{table}

%------------------------------------------------------------------------
\section{Conclusion}
% In this work, we made the first endeavour to tackle the challenging domain adaption person re-ID by one (few) shot learning. Different from the common practice of adopting the step-wise strategy to iteratively assign pseudo labels to unlabeled samples, we introduced a similarity-guided strategy that enables the network to ``see" all the samples of target domain starting from the training, leading to the parameters can be quickly adapted from the source domain to the target domain. Extensive experimental results demonstrated that the performance of our approach outperformed the state-of-the-arts by a large margin.
In this work, we proposed Self-similarity Grouping (SSG), which can mine the potential similarity existing in target dataset automatically by different appearance cues (from global to local) in an unsupervised manner, to tackle the challenging domain adaption in person re-ID. Furthermore, we introduce a clustering-guided semi-supervised approach upon proposed SSG to adopt traditional one shot learning method to person re-ID, which is an open set problem. Extensive experimental results demonstrate that the performance of our approach outperforms the state-of-the-arts by a large margin.
%------------------------------------------------------------------------

\noindent {\bf Acknowledgements:} This work is part supported by IBM-ILLINOIS Center for Cognitive Computing Systems Research (C3SR) and ARC DECRA DE190101315.
{\small
\bibliographystyle{ieee_fullname}
\bibliography{egbib}
}
\end{document}